\documentclass{article} 
\usepackage{iclr2020_conference,times}

\usepackage{amsmath,amsfonts,bm}

\def\eqref#1{equation~\ref{#1}}

\def\1{\bm{1}}

\DeclareMathAlphabet{\mathsfit}{\encodingdefault}{\sfdefault}{m}{sl}
\SetMathAlphabet{\mathsfit}{bold}{\encodingdefault}{\sfdefault}{bx}{n}

\usepackage{hyperref}
\usepackage{url}
\usepackage{booktabs}
\usepackage{graphicx}
\usepackage{subfigure}
\usepackage{wrapfig}

\newcommand{\domX}{\mathcal{X}}
\newcommand{\domY}{\mathcal{Y}}
\newcommand{\bert}{\texttt{BERT}}
\newcommand{\enc}{\texttt{Enc}}
\newcommand{\dec}{\texttt{Dec}}
\newcommand{\attn}{\texttt{attn}}
\newcommand{\ffn}{\texttt{FFN}}
\newcommand{\myeqref}[1]{Eqn.(\ref{#1})}

\title{Incorporating BERT into \\Neural Machine Translation}

\author{Jinhua Zhu$^{1,*}$, Yingce Xia$^{2,}$\thanks{This work is conducted at Microsoft Research Asia. The first two authors contributed equally to this work.}\,\,, Lijun Wu$^3$, ~Di He$^4$,\\
\textbf{Tao Qin}$^2$, \textbf{Wengang Zhou}$^1$, \textbf{Houqiang Li}$^1$, \textbf{Tie{-}Yan Liu}$^2$\\
$^1$CAS Key Laboratory of GIPAS, EEIS Department, University of Science and Technology of China;\\ $^2$Microsoft Research; \\
$^3$Sun Yat-sen University;\\
$^4$Key Laboratory of Machine Perception (MOE), School of EECS, Peking University\\
$^1$\texttt{teslazhu@mail.ustc.edu.cn},\;\texttt{\{zhwg,lihq\}@ustc.edu.cn}\\
$^2$\texttt{yingce.xia@gmail.com,\, \{taoqin,tyliu\}@microsoft.com}\\
$^3$\texttt{wulijun3@mail2.sysu.edu.cn}\;\;$^4$\texttt{di\_he@pku.edu.cn}
}

\iclrfinalcopy 
\begin{document}

\maketitle

\begin{abstract}
The recently proposed BERT~\citep{devlin2018bert} has shown great power on a variety of natural language understanding tasks, such as text classification, reading comprehension, etc. However, how to effectively apply BERT to neural machine translation (NMT) lacks enough exploration. While BERT is more commonly used as fine-tuning instead of contextual embedding for downstream language understanding tasks, in NMT, our preliminary exploration of using BERT as contextual embedding is better than using for fine-tuning. This motivates us to think how to better leverage BERT for NMT along this direction. We propose a new algorithm named BERT-fused model, in which we first use BERT to extract representations for an input sequence, and then the representations are fused with each layer of the encoder and decoder of the NMT model through attention mechanisms. We conduct experiments on supervised (including sentence-level and document-level translations), semi-supervised and unsupervised machine translation, and achieve state-of-the-art results on seven benchmark datasets. Our code is available at  \url{https://github.com/bert-nmt/bert-nmt}. 
\end{abstract}

\section{Introduction}
Recently, pre-training techniques, like ELMo~\citep{peters2018deep}, GPT/GPT-2~\citep{radford2018improving,radford2019language}, BERT~\citep{devlin2018bert}, cross-lingual language model (briefly, XLM)~\citep{lample2019cross}, XLNet~\citep{yang2019xlnet} and RoBERTa~\citep{liu2019roberta} have attracted more and more attention in machine learning and natural language processing communities. The models are first pre-trained on large amount of unlabeled data to capture rich representations of the input, and then applied to the downstream tasks by either providing context-aware embeddings of an input sequence~\citep{peters2018deep}, or initializing the parameters of the downstream model~\citep{devlin2018bert} for fine-tuning. Such pre-training approaches lead to significant improvements on natural language understanding tasks. Among them, BERT is one of the most powerful techniques that inspires lots of variants like XLNet, XLM, RoBERTa and achieves state-of-the-art results for many language understanding tasks including reading comprehension, text classification, etc \citep{devlin2018bert}.


Neural Machine Translation (NMT) aims to translate an input sequence from a source language to a target language. An NMT model usually consists of an encoder to map an input sequence to hidden representations, and a decoder to decode hidden representations to generate a sentence in the target language. Given that BERT has achieved great success in language understanding tasks, a question worthy studying is how to incorporate BERT to improve NMT. Due to the computation resource limitation, training a BERT model from scratch is unaffordable for many researchers. Thus, we focus on the setting of leveraging a pre-trained BERT model (instead of training a BERT model from scratch) for NMT.

Given that there is limited work leveraging BERT for NMT, our first attempt is to try two previous strategies: (1) using BERT to initialize downstream models and then fine-tuning the models, and (2) using BERT as context-aware embeddings for downstream models. For the first strategy, following~\cite{devlin2018bert}, we initialize the encoder of an NMT model with a pre-trained BERT model, and then finetune the NMT model on the downstream datasets. Unfortunately, we did not observe significant improvement. Using a pre-trained XLM~\citep{lample2019cross} model, a variant of BERT for machine translation, to warm up an NMT model is another choice. XLM has been verified to be helpful for WMT'16 Romanian-to-English translation. But when applied to a language domain beyond the corpus for training XLM (such as IWSLT dataset~\citep{iwslt14}, which is about spoken languages) or when large bilingual data is available for downstream tasks, no significant improvement is observed neither. For the second strategy, following the practice of ~\citep{peters2018deep}, we use BERT to provide context-aware embeddings for the NMT model. We find that this strategy outperforms the first one (please refer to Section~\ref{sec:init_explore} for more details). This motivates us to go along this direction and design more effective algorithms. 

We propose a new algorithm, {\em BERT-fused model}, in which we exploit the representation from BERT by feeding it into all layers rather than served as input embeddings only. We use the attention mechanism to adaptively control how each layer interacts with the representations, and deal with the case that BERT module and NMT module might use different word segmentation rules, resulting in different sequence (i.e., representation) lengths. Compared to standard NMT, in addition to BERT, there are two extra attention modules, the BERT-encoder attention and BERT-decoder attention. An input sequence is first transformed into representations processed by BERT. Then, by the BERT-encoder attention module, each NMT encoder layer interacts with the representations obtained from BERT and eventually outputs fused representations leveraging both BERT and the NMT encoder. The decoder works similarly and fuses BERT representations and NMT encoder representations. 

We conduct $14$ experiments on various NMT tasks to verify our approach, including supervised, semi-supervised and unsupervised settings. For supervised NMT, we work on five tasks of IWSLT datasets and two WMT datasets. Specifically, we achieve $36.11$ BLEU score on IWSLT'14 German-to-English translation, setting a new record on this task. We also work on two document-level translations of IWSLT, and further boost the BLEU score of German-to-English translation to $36.69$. On WMT'14 datasets, we achieve $30.75$ BLEU score on English-to-German translation and $43.78$ on English-to-French translation, significantly better over the baselines. For semi-supervised NMT, we boost BLEU scores of WMT'16 Romanian-to-English translation with back translation~\citep{sennrich2016improving}, a classic semi-supervised algorithm, from $37.73$ to $39.10$, achieving the best result on this task. Finally, we verify our algorithm on unsupervised English$\leftrightarrow$French and unsupervised English$\leftrightarrow$Romanian translations and also achieve state-of-the-art results. 

\section{Background and related work}
\label{sec:related_work}
We briefly introduce the background of NMT and review current pre-training techniques.

\noindent{\textbf{NMT}} aims to translate an input sentence from the source language to the target one. An NMT model usually consists of an encoder, a decoder and an attention module. The encoder maps the input sequence to hidden representations and the decoder maps the hidden representations to the target sequence. The attention module is first introduced by~\cite{bahdanau2014neural}, which is used to better align source words and target words. The encoder and decoder can be specialized as LSTM~\citep{Hochreiter:1997:LSM:1246443.1246450,sutskever2014sequence,wu2016google}, CNN~\citep{gehring2017convolutional} and Transformer~\citep{vaswani2017attention}. A Transformer layer consists of three sub-layers, a self-attention layer that processes sequential data taking the context of each timestep into consideration, an optional encoder-decoder attention layer that bridges the input sequence and target sequence which exists in decoder only, and a feed-forward layer for non-linear transformation. Transformer achieves the state-of-the-art results for NMT ~\citep{barrault-EtAl:2019:WMT}. In this work, we will use Transformer as the basic architecture of our model.

\noindent{\textbf{Pre-training}} has a long history in machine learning and natural language processing~\citep{erhan2009difficulty,erhan2010does}. \citet{mikolov2013distributed} and \citet{pennington2014glove} proposed to use distributional representations (i.e., word embeddings) for individual words. \cite{dai2015semi} proposed to train a language model or an auto-encoder with unlabeled data and then leveraged the obtained model to finetune downstream tasks. Pre-training has attracted more and more attention in recent years and achieved great improvements when the data scale becomes large and deep neural networks are employed. ELMo was proposed in \citet{peters2018deep} based on bidirectional LSTMs and its pre-trained models are fed into downstream tasks as context-aware inputs. In GPT~\citep{radford2018improving}, a Transformer based language model is pre-trained on unlabeled dataset and then finetuned on downstream tasks. BERT~\citep{devlin2018bert} is one of the widely adopted pre-training approach for model initialization. The architecture of BERT is the encoder of Transformer~\citep{vaswani2017attention}. Two kinds of objective functions are used in BERT training: (1) {\em Masked language modeling (MLM)}, where $15\%$ words in a sentence are masked and BERT is trained to predict them with their surrounding words. 
(2) {\em Next sentence prediction (NSP)}: Another task of pre-training BERT is to predict whether two input sequences are adjacent. For this purpose, the training corpus consists of tuples (\texttt{[cls]}, \texttt{input} 1, \texttt{[sep]}, \texttt{input} 2, \texttt{[sep]}), with learnable special tokens \texttt{[cls]} to classify whether \texttt{input} 1 and \texttt{input} 2 are adjacent and \texttt{[sep]} to segment two sentences, and with probability 50\%, the second input is replaced with a random input. Variants of BERT have been proposed: In XLM~\citep{lample2019cross}, the model is pre-trained based on multiple languages and \emph{NSP} task is removed; in RoBERTa~\citep{liu2019roberta}, more unlabeled data is leveraged without {\em NSP} task neither; in XLNet~\citep{yang2019xlnet}, a permutation based modeling is introduced. 

\section{A preliminary exploration}
\label{sec:init_explore}
While a few pieces of work \citep{lample2019cross,song2019mass} design specific pre-training methods for NMT, they are time and resource consuming given that they need to pre-train large models from scratch using large-scale data, and even one model for each language pair. In this work, we focus on the setting of using a pre-trained BERT model. Detailed model download links can be found in Appendix \ref{app:model_download}.

Considering that pre-trained models have been utilized in two different ways for other natural language tasks, it is straightforward to try them for NMT. Following previous practice, we make the following attempts.

\noindent(I) Use pre-trained models to initialize the NMT model. There are different implementations for this approach. (1) Following~\citep{devlin2018bert}, we initialize the encoder of an NMT model with a pre-trained BERT. (2) Following~\citep{lample2019cross}, we initialize the encoder and/or decoder of an NMT model with XLM. 

\noindent(II) Use pre-trained models as inputs to the NMT model. Inspired from~\citep{peters2018deep}, we feed the outputs of the last layer of BERT to an NMT model as its inputs.

We conduct experiments on the IWSLT'14 English$\to$German translation, a widely adopted dataset for machine translation consisting of $160k$ labeled sentence pairs. We choose Transformer~\citep{vaswani2017attention} as the basic model architecture with \texttt{transformer\_iwslt\_de\_en} configuration (a six-layer model with $36.7$M parameters). The translation quality is evaluated by BLEU~\citep{papineni2002bleu} score; the larger, the better. Both BERT$_{\text{base}}$ and XLM models are pre-trained and we get them from the Web. More details about the experimental settings are included in Appendix \ref{preliminary}.

\begin{table}[!htbp]
    \small
	\centering
	\caption{Preliminary explorations on IWSLT'14 English$\to$German translation.}
	\begin{tabular}{lcc}
		\toprule
		Algorithm & BLEU score  \\
		\midrule
		Standard Transformer  & $28.57$  \\
		\midrule
		Use BERT to initialize the encoder of NMT & $27.14$ \\
		Use XLM to initialize the encoder of NMT & $28.22$  \\
		Use XLM to initialize the decoder of NMT & $26.13$  \\
		Use XLM to initialize both the encoder and decoder of NMT & $28.99$ \\
		\midrule
		Leveraging the output of BERT as embeddings & $29.67$ \\
		\bottomrule
	\end{tabular}
	\label{tab:results_iwslt_first_explore}
\end{table}

The results are shown in Table~\ref{tab:results_iwslt_first_explore}. We have several observations: (1) Using BERT to initialize the encoder of NMT can only achieve $27.14$ BLEU score, which is even worse than standard Transformer without using BERT. That is, simply using BERT to warm up an NMT model is not a good choice. (2) Using XLM to initialize the encoder or decoder respectively, we get $28.22$ or $26.13$ BLEU score, which does not outperform the baseline. If both modules are initialized with XLM, the BLEU score is boosted to $28.99$, slightly outperforming the baseline. Although XLM achieved great success on WMT'16 Romanian-to-English, we get limited improvement here. Our conjecture is that the XLM model is pre-trained on news data, which is out-of-domain for IWSLT dataset mainly about spoken languages and thus, leading to limited improvement. (3) When using the output of BERT as context-aware embeddings of the encoder, we achieve $29.67$ BLEU, much better than using pre-trained models for initialization.  This shows that leveraging BERT as a feature provider is more effective in NMT. This motivates us to take one step further and study how to fully exploit such features provided by pre-trained BERT models.

\section{Algorithm}\label{sec:framework}
In this section, we first define the necessary notations, then introduce our proposed BERT-fused model and finally provide discussions with existing works. 

\noindent\textbf{Notations} Let $\domX$ and $\domY$ denote the source language domain and target language domain respectively, which are the collections of sentences with the corresponding languages. For any sentence $x\in\domX$ and $y\in\domY$, let $l_x$ and $l_y$ denote the number of units (e.g., words or sub-words) in $x$ and $y$. The $i$-th unit in $x$/$y$ is denoted as $x_i$/$y_i$. Denote the encoder, decoder and BERT as $\enc$, $\dec$ and $\bert$ respectively. For ease of reference, we call the encoder and decoder in our work as the {\em NMT module}. W.l.o.g., we assume both the encoder and decoder consists of $L$ layers. Let $\attn(q,K,V)$ denote the attention layer, where $q$, $K$ and $V$ indicate query, key and value respectively~\citep{vaswani2017attention}. We use the same feed-forward layer as that used in \citep{vaswani2017attention} and denote it as $\ffn$. Mathematical formulations of the above layers are left at Appendix~\ref{app:notation_details}.

\subsection{BERT-fused model}
\begin{figure}[!htbp]
\centering
\includegraphics[width=0.90\linewidth]{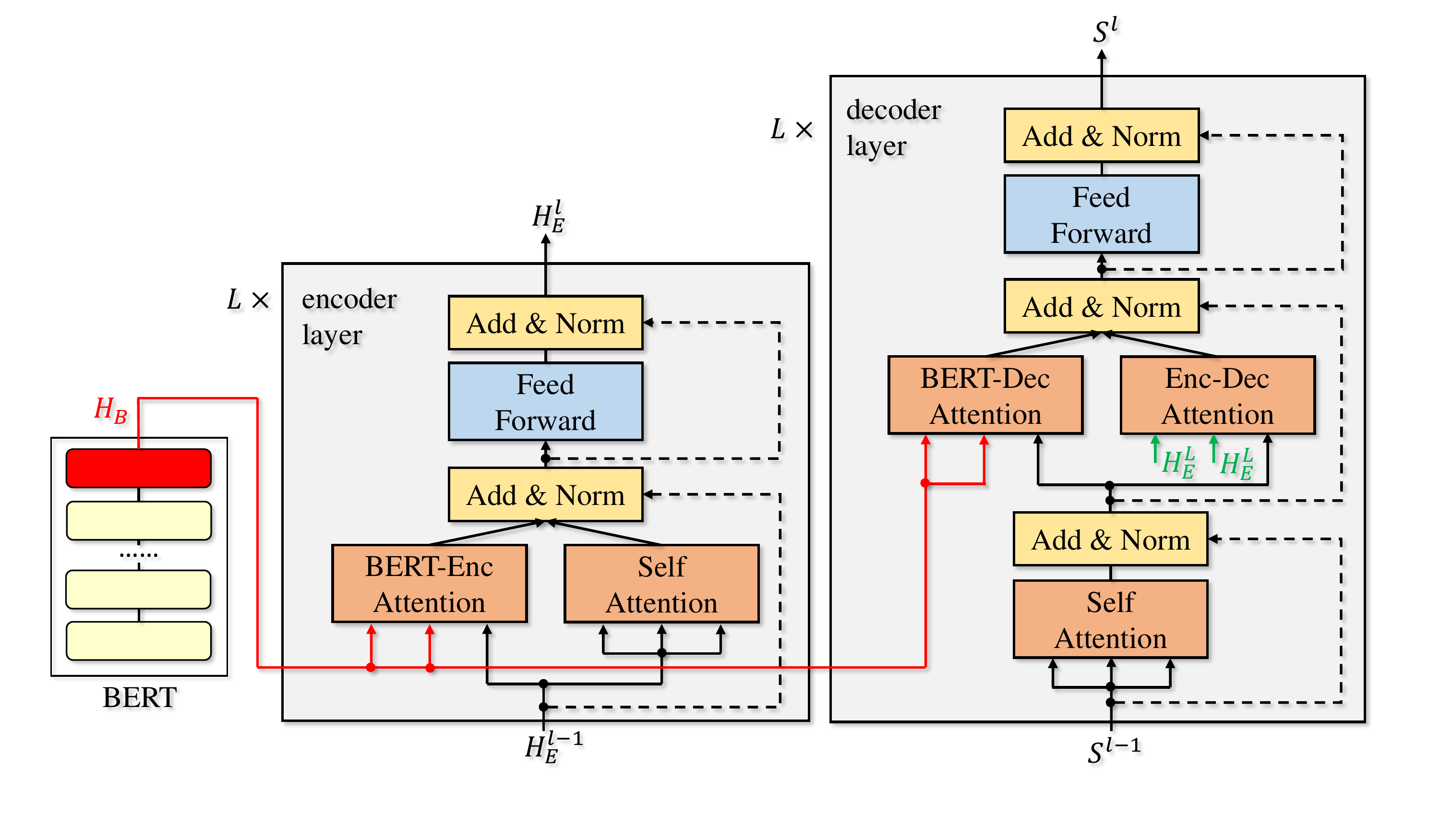}
\caption{The architecture of BERT-fused model. The left and right figures represent the BERT, encoder and decoder respectively. Dash lines denote residual connections. $H_B$ (red part) and $H^L_E$ (green part) denote the output of the last layer from BERT and encoder.}
\label{fig:framework}
\end{figure}

An illustration of our algorithm is shown in Figure~\ref{fig:framework}. Any input $x\in\domX$ is progressively processed by the BERT, encoder and decoder.

\noindent{\em Step-1}: Given any input $x\in\domX$, $\bert$ first encodes it into representation $H_B=\bert(x)$. $H_B$ is the output of the last layer in $\bert$. The $h_{B,i}\in H_B$ is the representation of the $i$-th wordpiece in $x$. 

\noindent{\em Step-2}: Let $H^l_E$ denote the hidden representation of $l$-th layer in the encoder, and let $H^0_E$ denote word embedding of sequence $x$. Denote the $i$-th element in $H^l_E$ as $h^l_i$ for any $i\in[l_x]$. In the $l$-th layer, $l\in[L]$,
\begin{equation}
\tilde{h}^l_i=\frac{1}{2}\big(\attn_S(h^{l-1}_i,H^{l-1}_E,H^{l-1}_E) + \attn_B(h^{l-1}_i,H_B,H_B)\big),\,\forall i\in[l_x],
\label{eq:encoder_module}
\end{equation}
where $\attn_S$ and $\attn_B$ are attention models (see Eqn.(\ref{eqn:attention_model})) with different parameters. Then each $\tilde{h}^l_i$ is further processed by $\ffn(\cdot)$ defined in \myeqref{eq:ffn} and we get the output of the $l$-th layer: $H^l_E=(\ffn(\tilde{h}^l_1),\cdots,\ffn(\tilde{h}^l_{l_x}))$. The encoder will eventually output $H^L_E$ from the last layer.

\noindent{\em Step-3}: Let $S^l_{<t}$ denote the hidden state of $l$-th layer in the decoder preceding time step $t$, i.e., $S^l_{<t}=(s^l_1,\cdots,s^l_{t-1})$. Note $s^0_1$ is a special token indicating the start of a sequence, and $s^0_t$ is the embedding of the predicted word at time-step $t-1$. At the $l$-th layer, we have
\begin{equation}
\begin{aligned}
&\hat{s}^l_{t}=\attn_S(s^{l-1}_t, S^{l-1}_{<t+1},S^{l-1}_{<t+1});\\
& \tilde{s}^l_{t}=\frac{1}{2}\big(\attn_B(\hat{s}^l_{t}, H_B,H_B)+\attn_E(\hat{s}^l_{t}, H_E^L,H_E^L)\big),\;
s^l_t = \ffn(\tilde{s}^l_t).
\end{aligned}
\label{eq:decoder_module}
\end{equation}
The $\attn_S$, $\attn_B$ and $\attn_E$ represent self-attention model, BERT-decoder attention model and encoder-decoder attention model respectively. \myeqref{eq:decoder_module} iterates over layers and we can eventually obtain $s^L_t$. Finally $s^L_t$ is mapped via a linear transformation and softmax to get the $t$-th predicted word $\hat{y}_t$. The decoding process continues until meeting the end-of-sentence token.

In our framework, the output of BERT serves as an external sequence representation, and we use an attention model to incorporate it into the NMT model. This is a general way to leverage the pre-trained model regardless of the tokenization way. 

\subsection{Drop-net trick}
Inspired by dropout~\citep{srivastava2014dropout} and drop-path~\citep{larsson2017fractalnet}, which can regularize the network training, we propose a drop-net trick to ensure that the features output by BERT and the conventional encoder are fully utilized. The drop-net will effect \myeqref{eq:encoder_module} and \myeqref{eq:decoder_module}. Denote the drop-net rate as $p_{\text{net}}\in[0,1]$. At each training iteration, for any layer $l$, we uniformly sample a random variable $U^l$ from $[0,1]$, then all the $\tilde{h}^l_i$ in \myeqref{eq:encoder_module} are calculated in the following way:
\begin{small}
\begin{equation}
\begin{aligned}
\tilde{h}^l_{i,\text{drop-net}}& =\mathbb{I}\big(U^l<\frac{p_{\text{net}}}{2}\big)\cdot\attn_S(h^{l-1}_i,H^{l-1}_E,H^{l-1}_E) + \mathbb{I}\big(U^l>1-\frac{p_{\text{net}}}{2}\big)\cdot\attn_B(h^{l-1}_i,H_B,H_B)\\
& + \frac{1}{2}\mathbb{I}\big(\frac{p_{\text{net}}}{2} \le U^l \le 1-\frac{p_{\text{net}}}{2}\big)\cdot\big(\attn_S(h^{l-1}_i,H^{l-1}_E,H^{l-1}_E) + \attn_B(h^{l-1}_i,H_B,H_B)\big),
\end{aligned}
\label{eq:encoder_module_dropnet_training}
\end{equation}
\end{small}
where $\mathbb{I}(\cdot)$ is the indicator function. For any layer, with probability $p_{\text{net}}/2$, either the BERT-encoder attention or self-attention is used only; w.p. $(1-p_{\text{net}})$, both the two attention models are used. For example, at a specific iteration, the first layer might uses $\attn_S$ only while the second layer uses $\attn_B$ only. During inference time, the expected output of each attention model is used, which is $\mathbb{E}_{U\sim\text{uniform}[0,1]}(\tilde{h}^l_{i,\text{drop-net}})$. The expectation is exactly \myeqref{eq:encoder_module}.

Similarly, for training of the decoder, with the drop-net trick, we have 
\begin{small}
\begin{equation}
\begin{aligned}
\tilde{s}^l_{t,\text{drop-net}}&=\mathbb{I}(U^l<\frac{p_{\text{net}}}{2})\cdot\attn_B(\hat{s}^l_{t}, H_B,H_B)+\mathbb{I}(U^l>1-\frac{p_{\text{net}}}{2})\cdot\attn_E(\hat{s}^l_{t}, H_E^L,H_E^L) \\
&+\frac{1}{2}\mathbb{I}(\frac{p_{\text{net}}}{2}\le U^l \le 1-\frac{p_{\text{net}}}{2})\cdot(\attn_B(\hat{s}^l_{t}, H_B,H_B)+\attn_E(\hat{s}^l_{t}, H_E^L,H_E^L)).
\end{aligned}
\label{eq:decoder_module_dropnet_training}
\end{equation}
\end{small}
For inference, it is calculated in the same way as \myeqref{eq:decoder_module}. Using this technique can prevent network from overfitting (see the second part of Section~\ref{sec:ablation_study} for more details).

\subsection{Discussion}

\noindent{\emph{Comparison with ELMo}} As introduced in Section~\ref{sec:related_work}, ELMo~\citep{peters2018deep} provides a context-aware embeddings for the encoder in order to capture richer information of the input sequence. Our approach is a more effective way of leveraging the features from the pre-trained model: (1) The output features of the pre-trained model are fused in all layers of the NMT module, ensuring the well-pre-trained features are fully exploited; (2) We use the attention model to bridge the NMT module and the pre-trained features of BERT, in which the NMT module can adaptively determine how to leverage the features from BERT.


\noindent{\emph{Limitations}} We are aware that our approach has several limitations. (1) Additional storage cost: our approach leverages a BERT model, which results in additional storage cost. However, considering the BLEU improvement and the fact that we do not need additional training of BERT, we believe that the additional storage is acceptable. (2) Additional inference time: We use BERT to encode the input sequence, which takes about $45\%$ additional time (see Appendix~\ref{sec:infer_time} for details). We will leave the improvement of the above two limitations as future work.

\section{Application to supervised NMT and semi-supervised NMT}\label{sec:supervised_nmt}
We first verify our BERT-fused model on the supervised setting, including low-resource and rich-resource scenarios. Then we conduct experiments on document-level translation to verify our approach. Finally, we combine BERT-fused model with back translation~\citep{sennrich2016improving} to verify the effectiveness of our method on semi-supervised NMT.

\subsection{Settings}\label{sec:sup_nmt:setting}
\noindent{\textbf{Dataset}} For the low-resource scenario, we choose IWSLT'14 English$\leftrightarrow$German (En$\leftrightarrow$De), English$\to$Spanish (En$\to$Es), IWSLT'17 English$\to$French (En$\to$Fr) and English$\to$Chinese (En$\to$Zh) translation. There are $160k$, $183k$, $236k$, $235k$ bilingual sentence pairs for En$\leftrightarrow$De, En$\to$Es, En$\to$Fr and En$\to$Zh tasks. Following the common practice~\citep{edunov2018classical}, for En$\leftrightarrow$De, we lowercase all words. All sentences are preprocessed by BPE~\citep{sennrich2016neural}. The model configuration is \texttt{transformer\_iwslt\_de\_en}, representing a six-layer model with embedding size $512$ and FFN layer dimension $1024$. For the rich-resource scenario, we work on WMT'14 En$\to$De and En$\to$Fr, whose corpus sizes are $4.5M$ and $36M$ respectively. We concatenate newstest2012 and newstest2013 as the validation set and use newstest2014 as the test set. The model configuration is \texttt{transformer\_big}, another six-layer network with embedding size $1024$ and FFN layer dimension $4096$.   More details about data and model  are left in Appendix~\ref{sec:iwslt_wmt_data}.

We choose BERT$_{\text{base}}$ for IWSLT tasks and BERT$_{\text{large}}$ for WMT tasks, which can ensure that the dimension of the BERT and NMT model almost match. The BERT models are fixed during training. Detailed BERT information for each task is in Appendix~\ref{app:pretrained_bert_models}. The drop-net rate $p_{\text{net}}$ is set as $1.0$.

\noindent{\textbf{Training Strategy}} We first train an NMT model until convergence, then initialize the encoder and decoder of the BERT-fused model  with the obtained model. The BERT-encoder attention and BERT-decoder attention are randomly initialized. Experiments on IWSLT and WMT tasks are conducted on $1$ and $8$ M40 GPUs respectively. The batchsize is $4k$ tokens per GPU. Following~\citep{ott2018scaling}, for WMT tasks, we accumulate the gradient for $16$ iterations and then update to simulate a $128$-GPU environment. It takes $1$, $8$ and $14$ days to obtain the pre-trained NMT models, and additional $1$, $7$ and $10$ days to finish the whole training process. The optimization algorithm is Adam~\citep{kingma2014adam} with initial learning rate $0.0005$ and \texttt{inverse\_sqrt} learning rate scheduler~\citep{vaswani2017attention}. For WMT'14 En$\to$De, we use beam search with width $4$ and length penalty $0.6$ for inference following~\citep{vaswani2017attention}. For other tasks, we use width $5$ and length penalty $1.0$.

\noindent{\textbf{Evaluation}} 
We use \texttt{multi-bleu.perl} to evaluate IWSLT'14 En$\leftrightarrow$De and WMT translation tasks for fair comparison with previous work. For the remaining tasks, we use a more advance implementation of BLEU score, \texttt{sacreBLEU} for evaluation. Script urls are in Appendix~\ref{sec:iwslt_wmt_data}.

\subsection{Results}

\begin{wraptable}{r}{5.5cm}
\vspace{-0.6cm}
\small
\centering
\caption{BLEU of all IWSLT  tasks.}
\begin{tabular}{c c c }
\toprule
& Transformer & BERT-fused  \\
\midrule
En$\to$De & $28.57$ & $30.45$ \\
De$\to$En & $34.64$ & $36.11$  \\
En$\to$Es & $39.0$ & $41.4$ \\
En$\to$Zh & $26.3$ & $28.2$ \\
En$\to$Fr & $35.9$ & $38.7$ \\
\bottomrule
\end{tabular}
\label{tab:results_iwslt_en-de}
\end{wraptable}

The results of IWSLT translation tasks are reported in Table~\ref{tab:results_iwslt_en-de}. We implemented standard Transformer as baseline. Our proposed BERT-fused model can improve the BLEU scores of the five tasks by $1.88$, $1.47$, $2.4$, $1.9$ and $2.8$ points respectively, demonstrating the effectiveness of our method. The consistent improvements on various tasks shows that our method works well for low-resource translations. We achieved state-of-the-art results on IWSLT'14 De$\to$En translation, a widely investigated baseline in machine translation. The comparison with previous methods are shown in Appendix \ref{app:iwsltdeen} due to space limitation.

The results of WMT'14 En$\to$De and En$\to$Fr are shown in Table~\ref{tab:results_wmt_en-de}. Our reproduced Transformer matches the results reported in \cite{ott2018scaling}, and we can see that our BERT-fused model can improve these two numbers to $30.75$ and $43.78$, achieving $1.63$ and $0.82$ points improvement. Our approach also outperforms the well-designed model DynamicConv~\citep{wu2019pay} and a model obtained through neural architecture search~\citep{so2019evolved}.

\begin{table}[!htbp]
\centering
\caption{BLEU scores of WMT'14 translation.}
\begin{tabular}{lcc}
\toprule
Algorithm & En$\to$De & En$\to$Fr \\
\midrule
DynamicConv~\citep{wu2019pay} & $29.7$ & $43.2$ \\
Evolved Transformer~\citep{so2019evolved} & $29.8$ & $41.3$ \\
\midrule
Transformer + Large Batch~\citep{ott2018scaling} & $29.3$ & $43.0$ \\
Our Reproduced  Transformer & $29.12$ & $42.96$ \\
Our BERT-fused model & $30.75$ & $43.78$ \\
\bottomrule
\end{tabular}
\label{tab:results_wmt_en-de}
\end{table}

\subsection{Translation with document-level contextual information}
BERT is able to capture the relation between two sentences, since the {\em next sentence prediction (NSP)} task is to predict whether two sentences are adjacent. We can leverage this property to improve translation with document-level contextual information~\citep{miculicich2018document}, which is briefly denoted as document-level translation. The inputs are a couple of sentences extracted from a paragraph/document, $x_1^d,x_2^d,\cdots,x_T^d$, where the $T$ $x$'s are contextually correlated. We want to translate them into target language by considering the contextual information.

\noindent{\textbf{Algorithm}} In our implementation, to translate a sentence $x$ to target domain, we leverage the contextual information by taking both $x$ and its preceding sentence $x_{\text{prev}}$ as inputs. $x$ is fed into $\enc$, which is the same as sentence-level translation. For the input of $\bert$, it is the concatenation of two sequences: (\texttt{[cls]}, $x_{\text{prev}}$, \texttt{[sep]}, $x$, \texttt{[sep]}), where both \texttt{[cls]} and \texttt{[sep]} are special tokens of $\bert$. 

\noindent{\textbf{Setting}} We use IWSLT'14 En$\leftrightarrow$De dataset as introduced in Section~\ref{sec:sup_nmt:setting}. The data is a collection of TED talks, where each talk consists of several sequences. We can extract the adjacent sentences for training, validation and test sets. The training strategy, hyperparameter selection and evaluation metric are the same for sentence-level translation. 

\begin{wraptable}{r}{7cm}
\vspace{-0.8cm}
\small
\centering
\caption{BLEU  of  document-level translation.}
\begin{tabular}{lcc}
\toprule
& En$\to$De & De$\to$En \\
\midrule
Sentence-level & $28.57$ & $34.64$ \\
\midrule
Our Document-level &$28.90$& $34.95$\\
\citet{miculicich2018document}&$27.94$&$33.97$ \\
\midrule
Sentence-level + BERT & $30.45$ & $36.11$ \\
Document-level + BERT & $31.02$ & $36.69$ \\
\bottomrule
\end{tabular}
\label{tab:results_iwslt_en-de_doc}
\vspace{-0.5cm}
\end{wraptable}

\noindent{\textbf{Baselines}} We use two baselines here. (1) To demonstrate how $\bert$ works in our model, we replace $\bert$ by a Transformer with configuration \texttt{transformer\_iwslt\_de\_en}, which is randomly initialized and jointly trained. (2) Another baseline is proposed by~\cite{miculicich2018document}, where multiple preceding sentences in a document are leveraged using a hierarchical attention network.

\noindent{\textbf{Results}} The results are shown in Table~\ref{tab:results_iwslt_en-de_doc}. We can see that introducing contextual information from an additional encoder can boost the sentence-level baselines, but the improvement is limited ($0.33$ for En$\to$De and $0.31$ for De$\to$En). For \citet{miculicich2018document}, the best results we obtain are $27.94$ and $33.97$ respectively, which are worse than the sentence-level baselines. Combining BERT-fused model and document-level information, we can eventually achieve $31.02$ for En$\to$De and $36.69$ for De$\to$En. 
We perform significant test\footnote{\url{https://github.com/moses-smt/mosesdecoder/blob/master/scripts/analysis/bootstrap-hypothesis-difference-significance.pl}} between sentence-level and document-level translation. Our document-level BERT-fused model significantly outperforms sentence-level baseline with $p$-value less than $0.01$. This shows that our approach not only works for sentence-level translation, but can also be generalized to document-level translation.

\subsection{Application to semi-supervised NMT}\label{sec:semisupervised_nmt}
We work on WMT'16 Romanian$\to$English (Ro$\to$En) translation to verify whether our approach can still make improvement over back translation~\citep{sennrich2016improving}, the standard and powerful semi-supervised way to leverage monolingual data in NMT.

The number of bilingual sentence pairs for Ro$\to$En is $0.6M$. \cite{sennrich2016edinburgh} provided $2M$ back translated data\footnote{Data at \url{http://data.statmt.org/rsennrich/wmt16_backtranslations/ro-en/}.}. We use newsdev2016 as validation set and newstest2016 as test set. Sentences were encoded using BPE with a shared source-target vocabulary of about 32$k$ tokens. We use \texttt{transformer\_big} configuration. Considering there is no Romanian BERT, we use the cased multilingual BERT (please refer to Appendix \ref{app:model_download}) to encode inputs. The drop-net rate $p_{\text{net}}$ is set as $1.0$. The translation quality is evaluated by  \texttt{multi-bleu.perl}.

\begin{wraptable}{r}{7cm}
\vspace{-0.8cm}
\centering
\caption{BLEU scores of WMT'16 Ro$\rightarrow$En.}
\begin{tabular}{lc}
\toprule
Methods& BLEU\\
\midrule
\citet{sennrich2016edinburgh}& $33.9$ \\
XLM~\citep{lample2019cross} & $38.5$ \\
\midrule
Standard Transformer & $33.12$  \\
+ back translation & $37.73$ \\
+ BERT-fused model & $39.10$ \\
\bottomrule
\end{tabular}
\label{tab:results_wmt_en-ro}
\vspace{-0.3cm}
\end{wraptable}

The results are shown in Table~\ref{tab:results_wmt_en-ro}. The Transformer baseline achieves $33.12$ BLEU score. With back-translation, the performance is boosted to $37.73$. We use the model obtained with back-translation to initialize BERT-fused model, and eventually reach $39.10$ BLEU. Such a score surpasses the previous best result $38.5$ achieved by XLM~\citep{lample2019cross} and sets a new record. This demonstrates that  our proposed approach is effective and can still achieve improvement over strong baselines.

\section{Ablation study}\label{sec:ablation_study}
We conduct two groups of ablation studies on IWSLT'14 En$\to$De translation to better understand our model.
\begin{table}[!htbp]
\centering
\caption{Ablation study on IWSLT'14 En$\to$De.}
\begin{tabular}{lc}
\toprule
Standard Transformer & 28.57\\
BERT-fused model & 30.45\\
\midrule
Randomly initialize encoder/decoder of BERT-fused model & $27.03$ \\
Jointly tune BERT and encoder/decoder of BERT-fused model& $28.87$ \\
\midrule
Feed BERT feature into all layers without attention & $29.61$ \\
Replace BERT output with random vectors & $28.91$ \\
Replace BERT with the encoder of another Transformer model& $28.99$\\ 
\midrule
Remove BERT-encoder attention & $29.87$ \\
Remove BERT-decoder attention & $29.90$ \\
\bottomrule
\end{tabular}
\label{tab:results_iwslt_en-de-ablation-study}
\end{table}

\noindent\textbf{ Study for training strategy and network architecture}

We conduct ablation study to investigate the performance of each component of our model and training strategy. Results are reported in Table~\ref{tab:results_iwslt_en-de-ablation-study}:

\noindent\textbf{(1)} We randomly initialize the NMT module (i.e., encoder and decoder) of BERT-fused model instead of using a warm-start one as introduced in the training strategy of Section~\ref{sec:sup_nmt:setting}. In this way, we can only achieve $27.03$ BLEU score, which cannot catch up with the baseline. We also jointly train BERT model with the NMT module. Although it can also boost the baseline from $28.57$ to $28.87$, it is not as good as fixing the BERT part, whose BLEU is $30.45$. 

\noindent\textbf{(2)} We feed the output of BERT into all layers of the encoder without attention models. That is, the \myeqref{eq:encoder_module} is revised to $\tilde{h}^l_i=\frac{1}{2}\big(\attn_S(h^{l-1}_i,H^{l-1}_E,H^{l-1}_E) + W^l_Bh^{l-1}_i)\big)$, where $W_B^l$ is learnable.
In this case, the encoder and BERT have to share the same vocabulary. The BLEU score is $29.61$, which is better than the standard Transformer but slightly worse than leveraging the output of BERT as embedding. This shows that the output of BERT should not be fused into each layer directly, and using the attention model to bridge the relation is better than using simple transformation. More results on different languages are included in Appendix \ref{app:feeding}.
To illustrate the effectiveness of our method, we choose another two kinds of ways to encode the input sequence rather than using BERT: (1) Using a fixed and randomly initialized embedding; (2) Using the encoder from another NMT model. Their BLEU scores are $28.91$ and $28.99$ respectively, indicating that the BERT pre-trained on large amount of unlabeled data can provide more helpful features to NMT.

\noindent\textbf{(3)} To verify where the output of BERT should be connected to, we remove the BERT-encoder attention (i.e., $\attn_B$ in \myeqref{eq:encoder_module}) and the BERT-decoder attention (i.e,, $\attn_B$ in \myeqref{eq:decoder_module}) respectively. Correspondingly, the BLEU score drops from $30.45$ to $29.87$ and $29.90$. This indicates that the output of BERT should be leveraged by both encoder and decoder to achieve better performances.
At last, considering that there are two stacked encoders in our model, we also choose ensemble models and deeper NMT models as baselines. Our approach outperforms the above baselines. The results are left in Appendix~\ref{app:more-ablation-study} due to space limitation.

\noindent\textbf{Study on drop-net}

\begin{figure}[!htpb]
\centering
\begin{minipage}{0.33\linewidth}
\subfigure[Training loss.]{
\includegraphics[width=\linewidth]{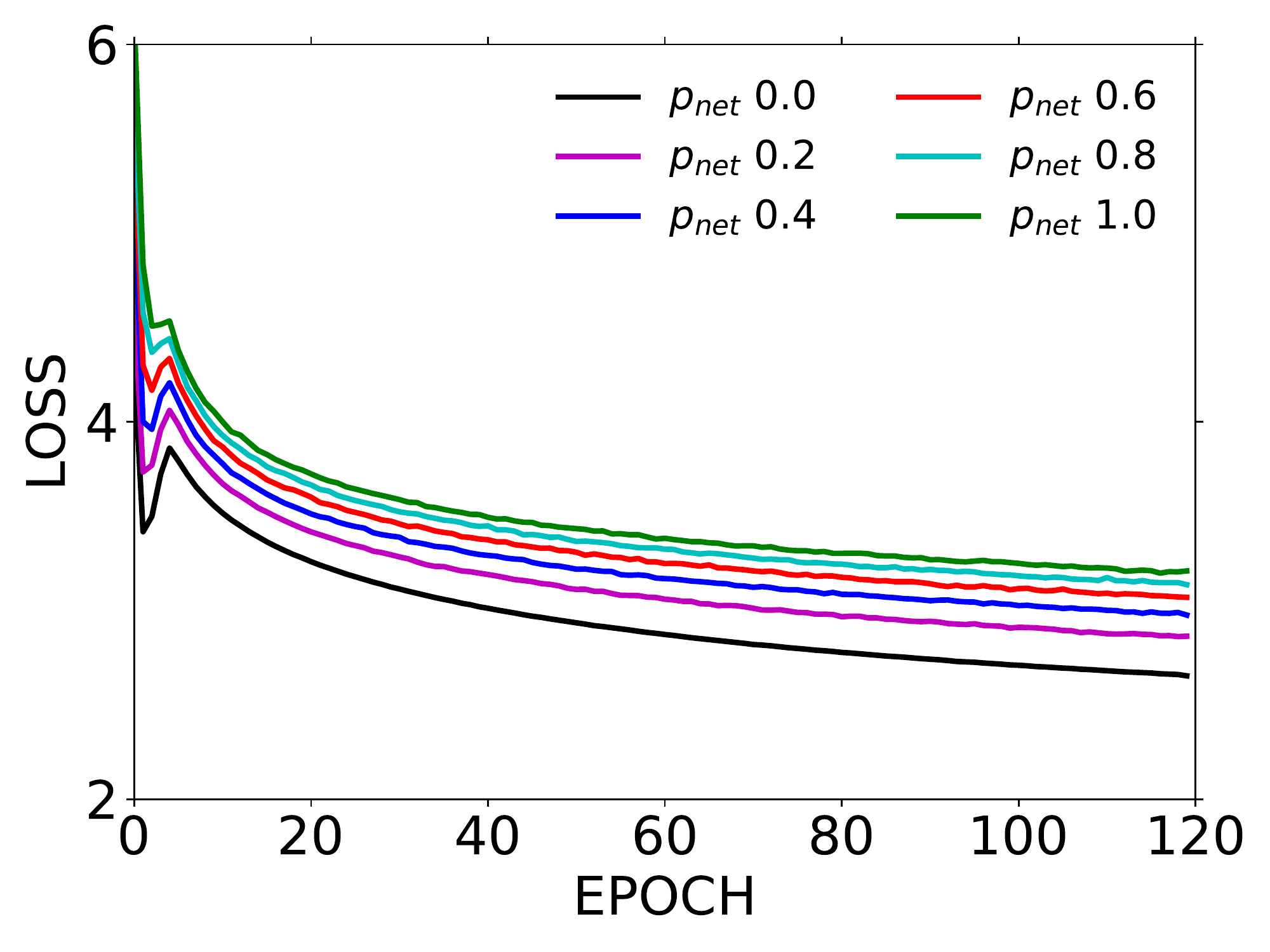}
}
\end{minipage}%
\begin{minipage}{0.33\linewidth}
\subfigure[Validation loss.]{
\includegraphics[width=\linewidth]{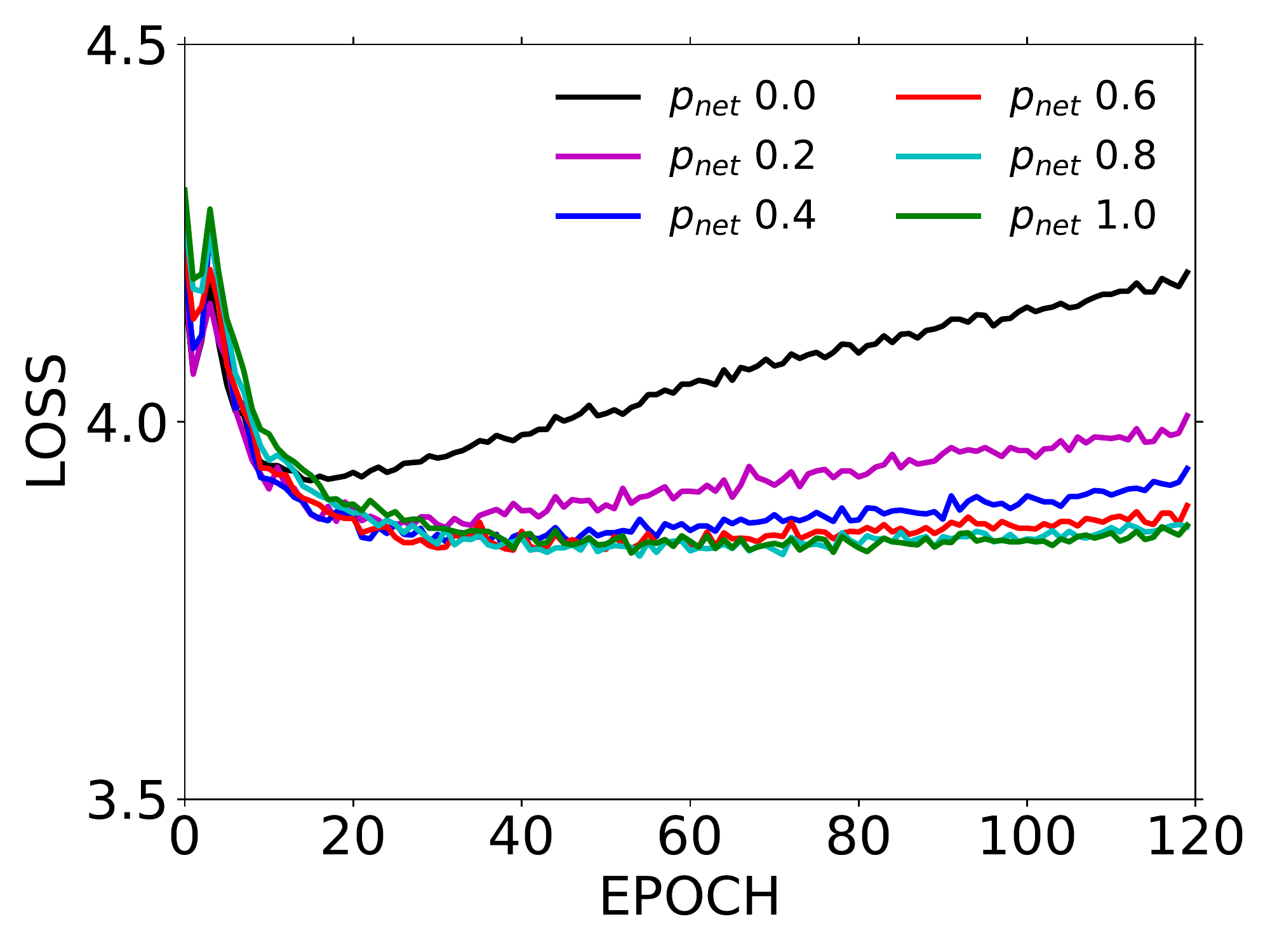}
}
\end{minipage}%
\begin{minipage}{0.33\linewidth}
\subfigure[Validation BLEU.]{
\includegraphics[width=\linewidth]{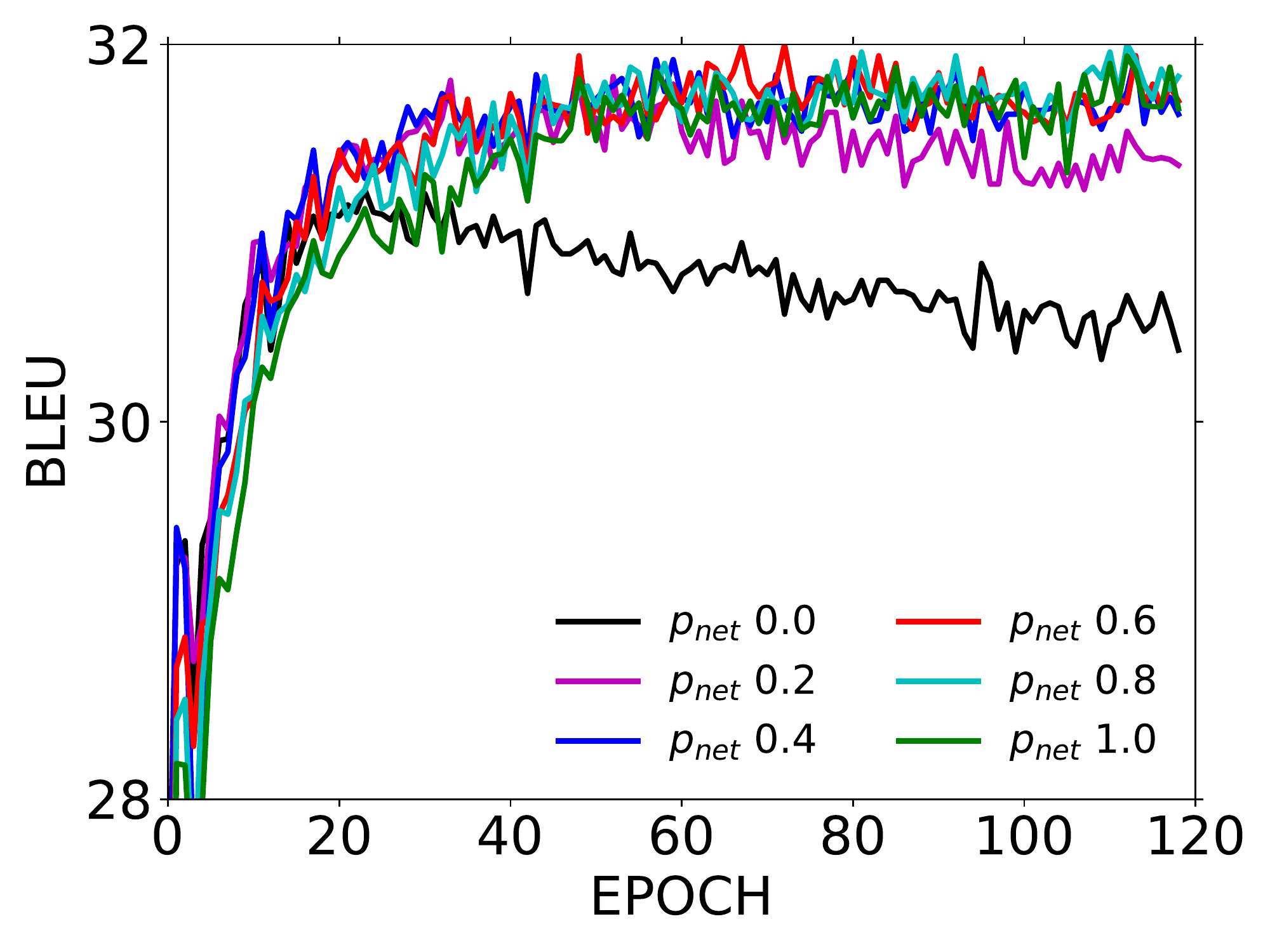}
}
\end{minipage}
\caption{Training/validation curves with different $p_{\text{net}}$'s.}
\label{fig:study_of_dropnet}
\end{figure}

To investigate the effect of drop-net, we  conduct experiments on IWSLT'14 En$\to$De dataset with different drop-net probability, $p_{\text{net}}\in\{0,0.2,0.4,0.6,0.8,1.0\}$. The results are shown in Figure~\ref{fig:study_of_dropnet}. As can been seen, although larger $p_{\text{net}}$ leads to larger training loss, it leads to smaller validation loss and so better BLUE scores. This shows that the drop-net trick can indeed improve the generalization ability of our model. We fix $p_{\text{net}}=1.0$ in other experiments unless specially specified. 

\section{Application to unsupervised NMT}\label{sec:unsupervised_nmt}
We work on unsupervised En$\leftrightarrow$Fr and En$\leftrightarrow$Ro translation. The data processing, architecture selection and training strategy is the same as \cite{lample2019cross}.

\noindent\textbf{Settings} For En$\leftrightarrow$Fr, we use $190M$ monolingual English sentences and $62M$ monolingual French sentences from WMT
News Crawl datasets, which is the same as that used in \citep{song2019mass}.\footnote{Data source: \url{https://modelrelease.blob.core.windows.net/mass/en-fr.tar.gz}.} For unsupervised En$\leftrightarrow$Ro translation, we use $50M$ English sentences from News Crawl (sampled from the data for En$\to$Fr) and collect $2.9M$ sentences for Romanian by concatenating News Crawl data sets and WMT'16 Romanian monolingual data following~\citet{lample2018phrase}. The data is preprocessed in the same way as~\citet{lample2019cross}. 

We use the same model configuration as \cite{lample2019cross}, with details in Appendix~\ref{app:unsup}. The $\bert$ is  the pre-trained XLM model (see Appendix~\ref{app:model_download}). We first train an unsupervised NMT model following~\citet{lample2019cross} until convergence. Then we initialize our BERT-fused  model with the obtained model and continue training. We train models on 8 M40 GPUs, and the batchsize is $2000$ tokens per GPU. We use the same optimization hyper-parameters as that described in \citet{lample2019cross}.

\begin{table}[!htbp]
\centering
\caption{BLEU scores of unsupervised NMT.}
\begin{tabular}{lcccc}
\toprule
& En$\to$Fr & Fr$\to$En & En$\to$Ro & Ro$\to$En \\
\midrule
\citet{lample2018phrase} & $27.6$&$27.7$& $25.1$& $23.9$\\
XLM~\citep{lample2019cross} & $33.4$ & $33.3$ &$33.3$&$31.8$\\
MASS~\citep{song2019mass} & $37.50$ & $34.90$ & $35.20$ & $33.10$ \\
Our BERT-fused model& $38.27$ & $35.62$  & $36.02$ & $33.20$\\
\bottomrule
\end{tabular}
\label{tab:results_unmt}
\end{table}

\noindent\textbf{Results} The results of unsupervised NMT are shown in Table~\ref{tab:results_unmt}.
With our proposed BERT-fused model, we can achieve $38.27$, $35.62$, $36.02$ and $33.20$ BLEU scores on the four tasks, setting state-of-the-art results on these tasks. Therefore,  our BERT-fused model also benefits unsupervised NMT. 

\section{Conclusion and future work}\label{sec:conc}
In this work, we propose an effective approach, BERT-fused model, to combine BERT and NMT, where the BERT is leveraged by the encoder and decoder through attention models. Experiments on supervised NMT (including sentence-level and document-level translations), semi-supervised NMT and unsupervised NMT demonstrate the effectiveness of our method. 

For future work, there are many interesting directions. First, we will study how to speed up the inference process. Second, we can apply such an algorithm to more applications, like questioning and answering. Third, how to compress BERT-fused model into a light version is another  topic. There are some contemporary works leveraging knowledge distillation to combine pre-trained models with NMT~\citep{yang2019towards,chen2019distilling}, which is a direction to explore.

\bibliographystyle{iclr2020_conference}
\bibliography{mybib}

\clearpage
\appendix
\section{Experiment Setup}

\subsection{IWSLT'14 \& WMT'14 Settings}\label{sec:iwslt_wmt_data}

We mainly follow the scripts below to preprocess the data:
{{\small
\url{https://github.com/pytorch/fairseq/tree/master/examples/translation} 
}}.

\noindent{\textbf{Dataset}} For the low-resource scenario, we choose IWSLT'14 English$\leftrightarrow$German (En$\leftrightarrow$De), English$\to$Spanish (En$\to$Es), IWSLT'17 English$\to$French (En$\to$Fr) and English$\to$Chinese (En$\to$Zh) translation. There are $160k$, $183k$, $236k$, $235k$ bilingual sentence pairs for En$\leftrightarrow$De, En$\to$Es, En$\to$Fr and En$\to$Zh tasks. Following the common practice~\citep{edunov2018classical}, for En$\leftrightarrow$De, we lowercase all words, split $7k$ sentence pairs from the training dataset for validation and concatenate \textit{dev2010}, \textit{dev2012}, \textit{tst2010}, \textit{tst2011}, \textit{tst2012} as the test set. For other tasks, we do not lowercase the words and use the official validation/test sets of the corresponding years. 

For rich-resource scenario, we work on WMT'14 En$\to$De and En$\to$Fr, whose corpus sizes are $4.5M$ and $36M$ respectively. We concatenate newstest2012 and newstest2013 as the validation set and use newstest2014 as the test set. 

We apply BPE~\citep{sennrich2016neural} to split words into sub-units. The numbers of BPE merge operation for IWSLT tasks, WMT'14 En$\to$De and En$\to$Fr are $10k$, $32k$ and $40k$ respectively. We merge the source and target language sentences for all tasks to build the vocabulary except En$\to$Zh. 

\noindent{\textbf{Model Configuration}} For IWSLT tasks, we use the \texttt{transformer\_iwslt\_de\_en} setting with dropout ratio $0.3$. In this setting, the embedding dimension, FFN layer dimension and number of layers are $512$, $1024$ and $6$. For WMT'14 En$\to$De and En$\to$Fr, we use \texttt{transformer\_big} setting (short for \texttt{transformer\_vaswani\_wmt\_en\_de\_big}) with dropout $0.3$ and $0.1$ respectively. In this setting, the aforementioned three parameters are $1024$, $4096$ and $6$ respectively. 

\noindent{\textbf{Evaluation}} 
We use \texttt{multi-bleu.perl}\footnote{\url{https://github.com/moses-smt/mosesdecoder/blob/master/scripts/generic/multi-bleu.perl}} to evaluate IWSLT'14 En$\leftrightarrow$De and WMT translation tasks for fair comparison with previous work. For the remaining tasks, we use a more advance implementation of BLEU score, detokenized sacreBLEU for evaluation\footnote{\url{https://github.com/mjpost/sacreBLEU}.}.

\subsection{Detailed Experiment Setting in Section 3}\label{preliminary}

The IWSLT'14 English-to-German data and model configuration is introduced in Section~\ref{sec:iwslt_wmt_data}.

For the training stategy, we use Adam~\citep{kingma2014adam} to optimize the network with $\beta_1=0.9$, $\beta_2=0.98$ and $\text{weight-decay}=0.0001$. The learning rate scheduler is \texttt{inverse\_sqrt}, where $\texttt{warmup-init-lr}=10^{-7}$, $\texttt{warmup-updates}=4000$ and $\texttt{max-lr}=0.0005$.

\subsection{Detailed model configuration in Unsupervised NMT}\label{app:unsup}
We leverage one Transformer model with GELU activation function to work on translations of two directions, where each language is associated with a language tag. The embedding dimension, FFN layer dimension and number of layer are $1024$, $4096$ and $6$. The $\bert$ is initialized by the pre-trained XLM model provided by~\citep{lample2019cross}. 

\section{More experiment results}

\subsection{More results on preliminary exploration of leveraging BERT}\label{app:more_results_xlm}
We use XLM to initialize the model for WMT'14 English$\to$German translation task, whose training corpus is relative large. We eventually obtain $28.09$ after $90$ epochs, which is still underperform the baseline, $29.12$ as we got. Similar problem is also reported in \url{https://github.com/facebookresearch/XLM/issues/32}. We leave the improvement of supervised NMT with XLM as future work.

\subsection{More ablation study}\label{app:more-ablation-study}

\noindent{\textbf{Part I: A different way to deal with multiple attention models}}

\cite{junczys2018ms} proposed a new way to handle multiple attention models. Instead of using \myeqref{eq:decoder_module}, the input is processed by self-attention, encoder-decoder attention and BERT-decoder attention sequentially. Formally,
\begin{equation}
\begin{aligned}
&\hat{s}^l_{t}=\attn_S(s^{l-1}_t, S^{l-1}_{<t+1},S^{l-1}_{<t+1});\\
&\bar{s}^l_t=\attn_E(\hat{s}^l_{t}, H_E^L,H_E^L);\\
& \tilde{s}^l_{t}=\attn_B(\bar{s}^l_{t}, H_B,H_B);\\
&s^l_t = \ffn(\tilde{s}^l_t).
\end{aligned}
\label{eq:decoder_module_stack}
\end{equation}
The BLEU score is $29.35$ for this setting, not as good as our proposed method.

\noindent{\textbf{Part II: More results on IWSLT'14 En$\to$De translation}}

Since our BERT-fused model contains two stacked encoders, we carry out two groups of additional baselines:

\noindent(1) Considering that stacking the BERT and encoder can be seen as a deeper model, we also train another two NMT models with deeper encoders, one with $18$ layers (since BERT$_\text{base}$ consists of $12$ layers) and the other with $12$ layers (which achieved best validation performance ranging from $6$ to $18$ layers). 

\noindent(2) We also compare the results of our approach with ensemble methods. To get an $M$-model ensemble, we independently train $M$ models with different random seeds ($M\in\mathbb{Z}_+$). We ensemble both standard Transformers and our BERT-fused  models, which are denoted as $M$-model ensemble (standard) and $M$-model ensemble (BERT-fused) respectively. Please note that when we aggregate multiple BERT-fused  models, we only need to store one replica of the BERT model because the BERT part is not optimized.

\begin{table}[!htbp]
\centering
\caption{More ablation study on IWSLT'14 En$\to$De.}
\begin{tabular}{lcc}
\toprule
Algorithm & BLEU  \\
\midrule
Standard Transformer & 28.57 \\
BERT-fused model & 30.45  \\
\midrule
$12$-layer encoder & $29.27$   \\
$18$-layer encoder & $28.92$  \\
\midrule
2-model ensemble (standard) & $29.71$  \\
3-model ensemble (standard) & $30.08$ \\
4-model ensemble (standard) & $30.18$ \\
\bottomrule
2-model ensemble (BERT-fused) & $31.09$ \\
3-model ensemble (BERT-fused) & $31.45$  \\
4-model ensemble (BERT-fused) & $31.85$ \\
\bottomrule
\end{tabular}
\label{tab:results_iwslt_en-de-ablation-more_params}
\end{table}

The results are shown in Table~\ref{tab:results_iwslt_en-de-ablation-more_params}. We have the following observations: 
\begin{enumerate}
\item Adding more layers can indeed boost the baseline, but still not as good as BERT-fused model. According to our experiments, when increasing the number of layers to $12$, we achieve the best BLEU score, $29.27$. 
\item We also compare our results to ensemble methods. Indeed, ensemble significantly boost the baseline by more than one point. However, even if using ensemble of four models, the BLEU score is still lower than our BERT-fused model (30.18 v.s. 30.45), which shows the effectiveness of our method. 
\end{enumerate}

We want to point out that our method is intrinsically different from ensemble. Ensemble approaches usually refer to ``independently'' train several different models for the same task, and then aggregate the output of each model to get the eventually task. In BERT-fused model, although we include a pre-trained BERT into our model, there is still only one model serving for the translation task.

In this sense, we can also combine our BERT-fused model with ensemble. Our approach benefits from ensemble too. When ensembling two models, we can achieve $31.09$ BLEU score. When adding the number of models to four, we eventually achieve $31.85$ BLEU score, which is $1.67$ point improvement over the ensemble of standard Transformer. 

\noindent{\textbf{Part III: More results on IWSLT'14 De$\to$En translation}}

We report the ensemble results on IWSLT'14 De$\to$En translation in Table~\ref{tab:results_iwslt_de2en-ablation-more_params}. We can get similar conclusion compared to that of IWSLT'14 En$\to$De. 

\begin{table}[!htbp]
\centering
\caption{More ablation study on IWSLT'14 De$\to$En.}
\begin{tabular}{lc}
\toprule
Algorithm & BLEU  \\
\midrule
Standard Transformer & $34.67$ \\
BERT-fused model & $36.11$ \\
\midrule
2-model ensemble (standard) & $35.92$  \\
3-model ensemble (standard) & $36.40$  \\
4-model ensemble (standard) & $36.54$  \\
\bottomrule
2-model ensemble (BERT-fused) & $37.42$  \\
3-model ensemble (BERT-fused) & $37.70$  \\
4-model ensemble (BERT-fused) & $37.71$  \\
\bottomrule
\end{tabular}
\label{tab:results_iwslt_de2en-ablation-more_params}
\end{table}

\subsection{More results on feeding BERT output to NMT module}\label{app:feeding}
The ablation study on more languages is shown in Table~\ref{tab:more_lang_ablation}. Our method achieves the best results compared to all baselines. 
\begin{table}[!htbp]
\centering
\caption{BLEU scores of IWSLT translation tasks.}
\begin{tabular}{lcccccc}
\toprule
Algorithm & En$\to$De & De$\to$En & En$\to$Es & En$\to$Zh &En$\to$Fr \\
\midrule
Standard Transformer & $28.57$ & $34.64$ &$39.0$ & $26.3$& $35.9$\\
Feed BERT feature into embedding & $29.67$ & $34.90$ & $39.5$ & $28.1$ & $37.3$  \\
Feed BERT feature into all layers of encoder &  $29.61$ & $34.84$ & $39.9$ & $28.1$ & $37.4$\\
Our BERT-fused model & $30.45$ & $36.11$ &$41.4$&$28.2$&$38.7$ \\
\bottomrule
\end{tabular}
\label{tab:more_lang_ablation}
\end{table}

\subsection{More baselines of IWSLT'14 German-to-English translation}\label{app:iwsltdeen}
We summarize the BLEU scores on IWSLT'14 De$\to$En of existed works and our BERT-fused model approach in Table~\ref{tab:summary:de2en:iwslt}. 

\begin{table}[!htbp]
\centering
\caption{Previous results of IWSLT'14 De$\to$En.}
\begin{tabular}{lc}
\toprule
Approach & BLEU \\ 
\midrule
Multi-agent dual learning~\citep{wang2018multi}  &  $35.56$ \\
Tied-Transformer~\citep{xia2019tied}  & $35.52$  \\
Loss to teach~\citep{wu2018learning} & $34.80$ \\
Role-interactive layer~\citep{weissenborn2019contextualized} & $34.74$ \\
Variational attention~\citep{deng2018latent} & $33.68$ \\
\midrule
Our BERT-fused model & $36.11$ \\
\bottomrule
\end{tabular}
\label{tab:summary:de2en:iwslt}
\end{table}

\subsection{Comparison with back translation}
When using unlabeled data to boost machine learning systems, one of the most notable approaches is back translation (briefly, BT)~\citep{sennrich2016improving}: We first train a reversed translation model, use the obtained model to translate the unlabeled data in the target domain back to source domain, obtain a synthetic dataset where the source data is back-translated and finally train the forward model on the augmented dataset. 

Our method has two main differences with BT method. 
\begin{enumerate}
\item In BT, the monolingual data from the target side is leveraged. In our proposed approach, we use a BERT of the source language, which indirectly leverages the monolingual data from the source side. In this way, our approach and BT are complementary to each other. In Section~\ref{sec:semisupervised_nmt}, we have already verified that our method can further improve the results of standard BT on Romanian-to-English translation.
\item To use BT, we have to train a reversed translation model and then back translate the monolingual data, which is time-cost due to the decoding process. In BERT-fused model, we only need to download a pre-trained BERT model, incorporate it into our model and continue training. Besides, the BERT module is fixed during training.
\end{enumerate}

On IWSLT'14, we also implement BT on wikipedia data, which is a subset of the corpus of training BERT. The model used for back translation are standard Transformer baselines introduced in Section~\ref{sec:supervised_nmt}, whose BLEU scores are $28.57$ and $34.64$ respectively. We back translate $1$M, $2$M, $5$M, $15$M and $25$M randomly selected German sentences. 

The results are reported in Table~\ref{tab:results_bt_wiki}. The rows started with BT($\cdot$) represent the results of BT, and the numbers in the brackets are the number of sentences for back translation. 


\begin{table}[!htbp]
\centering
\caption{BLEU scores IWSLT'14 En$\leftarrow$De by BT.}
\begin{tabular}{lc}
\toprule
Algorithm & En$\to$De  \\
\midrule
Standard Transformer & $28.57$ \\
BERT-fused model & $30.45$  \\
\midrule
BT (1M)  & $29.42$ \\
BT (2M)  & $29.76$ \\
BT (5M)  & $29.10$\\
BT (15M) & $28.26$ \\
BT (25M) & $27.34$ \\
\bottomrule
\end{tabular}
\label{tab:results_bt_wiki}
\end{table}


IWSLT dataset is a collection of spoken language, and the bilingual training corpus is small ($160k$). In Wikipedia, the sentences are relatively formal compared to the spoken language, which is  out-of-domain of spoken languages. We can see that when using $1$M or $2$M monolingual data for BT, the BLEU scores can indeed improve from $28.57$ to $29.42$/$29.76$. However, simply adding more wikipedia data for BT does not result in more improvement. There is even a slight drop when adding more than $15$M monolingual sentences. However, our BERT-fused model can achieve better performances than BT with wikipedia data.

\section{Comparison of inference time}\label{sec:infer_time}
\begin{table}[!htbp]
\centering
\caption{Comparisons on inference time (seconds), `+' is the increased ratio of inference time.}
\begin{tabular}{lccc}
\toprule
Dataset & Transformer & Ours & (+) \\
\midrule
IWSLT'14 En$\to$De & $70$ & $97$ & $38.6\%$ \\
IWSLT'14 De$\to$En & $69$ & $103$ & $49.3\%$ \\
WMT'14 En$\to$De   & $67$ & $99$  & $47.8\%$ \\
WMT'14 En$\to$Fr   & $89$ & $128$ & $43.8\%$ \\
\bottomrule
\end{tabular}
\label{tab:statistics_compare_bert_and_nmt}
\end{table}
We compare the inference time of our approach to the baselines. The results are shown in Table~\ref{tab:statistics_compare_bert_and_nmt},
where from the second column to the last column, the numbers are the inference time of standard Transformer, BERT-fused model, and the increase of inference time.

Indeed, introducing BERT to encode the input brings additional inference time, resulting in about 40\% to 49\% increase. But considering the significant improvement of BLEU score, it is acceptable of such extra cost. We will study how to reduce inference time in the future. 

\section{Download link of pre-trained BERT models}\label{app:pretrained_bert_models}
\label{app:model_download}
We leverage the pre-trained models provided by PyTorch-Transformers\footnote{\url{https://github.com/huggingface/pytorch-transformers}}.

For IWSLT'14 tasks, we choose BERT$_{\text{base}}$ model with $12$ layers and hidden dimension $768$.
\begin{enumerate}
\item IWSLT’14 En$\to$\{De, Es, Fr, Zh\}, we choose \texttt{bert-base-uncased}.
\item IWSLT’14 De$\to$En, we choose \texttt{bert-base-german-cased}.
\end{enumerate}
For WMT’14 En$\to$\{Fr, De\}, we choose \texttt{bert-large-uncased}, which is a BERT$_{\text{large}}$ model with $24$ layers and hidden dimension $1024$.

For WMT’16 Ro$\to$En, we choose \texttt{bert-base-multilingual-cased}, because there is no BERT specially trained for the Romanian. 

For unsupervised En$\leftrightarrow$Fr and unsupervised En$\leftrightarrow$Ro, we choose \texttt{xlm-mlm-enfr1024} and \texttt{xlm-mlm-enro1024} respectively.

The download links are summarized as follows:
\begin{itemize}
    \item bert-base-uncased: \url{https://s3.amazonaws.com/models.huggingface.co/bert/bert-base-uncased.tar.gz}.
    \item bert-large-uncased: \url{https://s3.amazonaws.com/models.huggingface.co/bert/bert-large-uncased.tar.gz}.
    \item bert-base-multilingual-cased: \url{https://s3.amazonaws.com/models.huggingface.co/bert/bert-base-multilingual-cased.tar.gz}.
    \item bert-base-german-cased: \url{https://int-deepset-models-bert.s3.eu-central-1.amazonaws.com/pytorch/bert-base-german-cased.tar.gz}.
    \item xlm-mlm-enfr\-1024: \url{https://s3.amazonaws.com/models.huggingface.co/bert/xlm-mlm-enfr-1024-pytorch_model.bin}.
    \item xlm-mlm-enro\-1024: \url{https://s3.amazonaws.com/models.huggingface.co/bert/xlm-mlm-enro-1024-pytorch_model.bin}.
\end{itemize}

\section{Details of the notations}\label{app:notation_details}
Let $\attn(q,K,V)$ denote the attention layer, where $q$, $K$ and $V$ indicate query, key and value respectively. Here $q$ is a $d_q$-dimensional vector ($d\in\mathbb{Z}$), $K$ and $V$ are two sets with $\vert K\vert=\vert V\vert$. Each $k_i\in K$ and $v_i\in V$ are also $d_k$/$d_v$-dimensional ($d_q$, $d_k$ and $d_v$ can be different) vectors, $i\in[\vert K\vert]$. The attention model works as follows:
\begin{equation}
\attn(q,K,V)=\sum_{i=1}^{\vert V\vert}\alpha_i
W_vv_i,\,\alpha_i=\frac{\exp\big((W_qq)^T(W_kk_i)\big)}{Z},\,Z=\sum_{i=1}^{\vert K\vert}\exp((W_qq)^T(W_kk_i)),
\label{eqn:attention_model}
\end{equation}
where $W_q$, $W_k$ and $W_v$ are the parameters to be learned. In \citet{vaswani2017attention}, $\attn$ is implemented as a multi-head attention model and we omit the details here to increase readability. Following~\citet{vaswani2017attention}, we define the non-linear transformation layer as 
\begin{equation}
\ffn(x)=W_2\max(W_1x+b_1,0)+b_2,
\label{eq:ffn}
\end{equation}
where $x$ is the input; $W_1$, $W_2$, $b_1$, $b_2$ are the parameters to be learned; $\max$ is an element-wise operator. Layer normalization is also applied following Transformer~\citep{vaswani2017attention}.

\end{document}